\documentclass[journal]{IEEEtran}
\usepackage{amsmath}
\usepackage{algorithmic}
\usepackage{array}

\usepackage{threeparttable}
\usepackage{tabularx}
\usepackage{color}
\usepackage{booktabs}
\usepackage{verbatim}
\usepackage{amsopn}
\usepackage{amsmath}
\usepackage{multirow}
\usepackage{colortbl}
\usepackage{graphicx}

\usepackage{hyperref}
\hypersetup{
colorlinks=true,
linkcolor=red,
filecolor=magenta,      
urlcolor=blue,
}

\usepackage{amsfonts}
\usepackage{bbm}
\usepackage{bm}

\usepackage{xspace}
\usepackage[mathscr]{euscript}
\usepackage{amssymb}
\makeatletter
\DeclareRobustCommand\onedot{\futurelet\@let@token\@onedot}
\def\@onedot{\ifx\@let@token.\else.\null\fi\xspace}

\def\eg{\emph{e.g}\onedot} 
\def\ie{\emph{i.e}\onedot}

\def\etal{\emph{et al}\onedot}
\makeatother

\begin{document}

\title{
Dual-Path Temporal Map Optimization for Make-up Temporal Video Grounding
}

\author{Jiaxiu Li$^\dag$, 
        Kun Li$^\dag$,
        Jia Li$^*$, 
        Guoliang Chen, 
        Dan Guo$^*$,
        and~Meng Wang,~\IEEEmembership{Fellow,~IEEE}
\thanks{
}
\thanks{
J. Li, K. Li, J. Li, G. Chen are with School of Computer Science and Information Engineering, Hefei University of Technology (HFUT), Hefei, 230601, China. (e-mail: jiaxiuli@mail.hfut.edu.cn, kunli.hfut@gmail.com, jiali@hfut.edu.cn, guoliangchen.hfut@gmail.com)

D. Guo and M. Wang are with School of Computer Science and Information Engineering, Hefei University of Technology (HFUT), Hefei, 230601, China, and are with Institute of Artificial Intelligence, Hefei Comprehensive National Science Center, Hefei, 230026, China (e-mail: guodan@hfut.edu.cn, eric.mengwang@gmail.com).

\IEEEcompsocthanksitem $^\dag$ Equal contribution. $^*$ Corresponding authors. 
}
}

\markboth{Journal of \LaTeX\ Class Files,~Vol.~14, No.~8, August~2015}%
{Shell \MakeLowercase{\textit{et al.}}: Bare Demo of IEEEtran.cls for IEEE Journals}

\maketitle

\begin{abstract}
Make-up temporal video grounding (MTVG) aims to localize the target video segment which is semantically related to a sentence describing a make-up activity, given a long video. Compared with the general video grounding task, MTVG focuses on meticulous actions and changes on the face. The make-up instruction step, usually involving detailed differences in products and facial areas, is more fine-grained than general activities (\eg, cooking activity and furniture assembly). Thus, existing general approaches cannot locate the target activity effectually. More specifically, existing proposal generation modules are not yet fully developed in providing semantic cues for the more fine-grained make-up semantic comprehension. To tackle this issue, we propose an effective proposal-based framework named \textit{Dual-Path Temporal Map Optimization Network (DPTMO)} to capture fine-grained multimodal semantic details of make-up activities. DPTMO extracts both query-agnostic and query-guided features to construct two proposal sets and uses specific evaluation methods for the two sets. Different from the commonly used single structure in previous methods, our dual-path structure can mine more semantic information in make-up videos and distinguish fine-grained actions well. These two candidate sets represent the cross-modal makeup video-text similarity and multi-modal fusion relationship, complementing each other. Each set corresponds to its respective optimization perspective, and their joint prediction enhances the accuracy of video timestamp prediction. Comprehensive experiments on the YouMakeup dataset demonstrate our proposed dual structure excels in fine-grained semantic comprehension. 
\end{abstract}


%
\IEEEpeerreviewmaketitle

\section{Introduction}
\IEEEPARstart{V}{ideo} understanding has gained increasing attention in the artificial intelligence field and developed diverse branches, including action detection~\cite{wu2020context,rana2021we}, video summarization~\cite{song2015tvsum,chu2015video,liu2020violin}, and video grounding~\cite{gao2017tall,mun2020local,li2021proposal}. 
Compared with common image recognition tasks, video understanding is typically more challenging and contains more semantic information and various actions. 
Among these branches, video grounding is a multi-modal understanding task that aims to retrieve a specific video moment described by a given query sentence from an untrimmed video.  
This task requires the model to understand the content of both complex textual queries and untrimmed videos.  

\begin{figure}[t]
\centering
\includegraphics[width=0.5\textwidth]{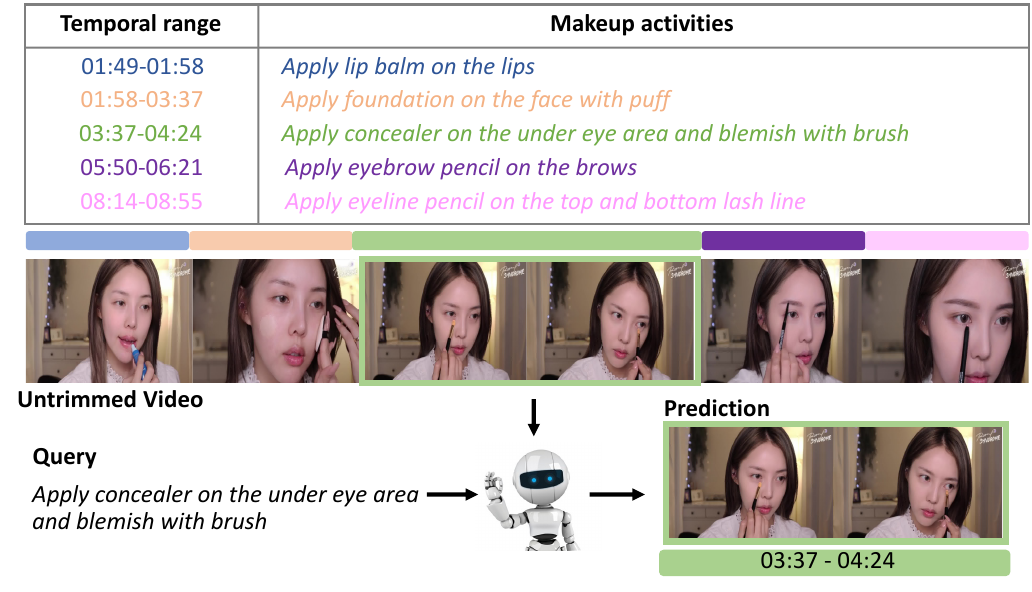}
\caption{An illustrative example of Make-up Temporal Video Grounding (MTVG). Given a video and a query, MTVG aims to localize the video segment corresponding to the query semantically.
}
\label{fig:task}
\end{figure}

Video grounding is also known as temporal sentence grounding in video~\cite{liu2021context,chen2018temporally,yuan2019find,li2023vigt}, video moment retrieval~\cite{gao2021fast,miech2021thinking,zeng2021multi} or natural language moment localization~\cite{liu2020jointly,ge2019mac,gao2017tall}. 
It has drawn considerable interest for its various potential applications, such as video entertainment~\cite{sanchez2015mood,ou2020multimodal}, robotic navigation~\cite{tellex2011understanding,gapon2021multimodal}, autonomous driving~\cite{caesar2020nuscenes,chiu2021probabilistic}. 
Existing methods can be broadly categorized into two groups, proposal-free methods~\cite{yuan2019find,zeng2020dense,chen2020rethinking,he2019read,wang2019language} and proposal-based methods~\cite{anne2017localizing,zhang2020learning,wang2022negative}. 
Proposal-free methods directly predict the start and end timestamps of the target video segment, while proposal-based methods firstly generate different candidate proposals and then rank them based on the similarity between query and proposal features. 
Early proposal-based works~\cite{anne2017localizing,gao2017tall} use sliding window to generate dense proposals, causing redundant computational costs. 
To avoid the above heavy computation, researchers propose to generate query-conditioned proposals~\cite{xu2019multilevel,chen2019semantic} and successfully reduce the number of candidates proposals. 
The performances of these proposal-generated methods largely depend on the quality of sampling candidates. 
However, most of them analyze proposals individually and ignore the importance of temporal dependencies. 
To address this issue, Zhang~\etal~\cite{zhang2020learning} propose a novel two-dimensional temporal map (2D Map) to enumerate all candidates with any length, and model temporal adjacent relations between proposals. 
Each row and column in the 2D map represents the video segment's start and end moment. 
Following 2D Map, Wang~\etal~\cite{wang2022negative} propose a metric-learning framework, which leverages the cross-modal mutual matching to give more supervision signals.

\begin{figure}[t]
\centering
\includegraphics[width=0.43\textwidth]{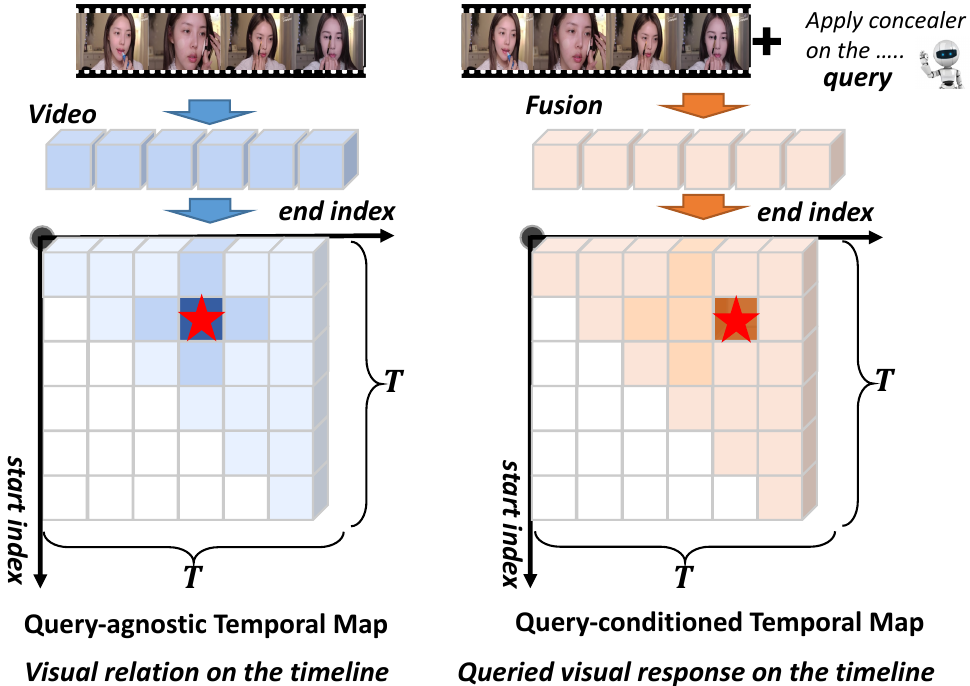}
\caption{The illustration of the construction of the query-agnostic and query-conditioned temporal maps. 
The query-agnostic map constructed on video features only can reflect different actions (\eg, apply eye on the brows, apply concealer on the under eye), while the query-conditioned map can reflect semantically related actions (can refer to the distribution of 2D maps). }
\label{fig:intro}
\end{figure}

The aforementioned works primarily focus on relatively distinguishable objects or activities, such as indoor or cooking activities. 
However, there is still significant room for improvement in these models when it comes to differentiating fine-grained semantic information.
In this paper, we focus on the fine-grained make-up domain. 
Compared to videos with drastic action changes in common video grounding tasks, make-up videos contain more fine-grained semantic information.
Additionally, these videos are longer than most of the existing instructional datasets, further complicating the task. 
Although previous methods can capture the general information in video content, they are inadequate for understanding the fine-grained semantics in the entire video. 
Therefore, how to understand the details of actions is the key to addressing MTVG. 
The reasons why we choose the make-up field are as follows: 
Firstly, the make-up video is more complex and challenging than other domains (\eg, cooking or indoor activities).
There are plenty of diverse steps in a specific make-up action. 
As shown in Fig.~\ref{fig:task}, the whole make-up process consists of several similar steps, such as \textit{``apply lip balm''}, \textit{``apply foundation''}, \textit{``apply eyeline pencil''}. 
In particular, all frames in the video share the same background, and the differences between each step are very subtle. 
These characteristics pose a challenge for the model to semantically match query with the correct video segments. 
The model is required to understand \textit{``apply concealer on the under eye area and blemish with brush''} and distinguish the semantic-matched segment from all similar steps. 
Secondly, make-up is closely related to our daily life and has significant commercial value. 
Make-up video analysis can greatly assist cosmetic companies and users in both editing and searching make-up processes efficiently.  
However, as the amount of video increase rapidly, manual make-up video retrieving has become labor-intensive and expensive, especially for long-time video data. 
Therefore, it is necessary to build a make-up video grounding video system that can automatically locate the target moment by language query. 

Based on the above considerations, we propose a novel proposal-based framework named \textit{Dual-Path Temporal Map Optimization Network (DPTMO)} illustrated in Fig.~\ref{fig:method} to model fine-grained semantics between the video and language query through two different 2D temporal maps. 
As shown in Fig.~\ref{fig:intro}, we implement 2D temporal map from two perspectives, \ie, query-agnostic and query-conditioned maps. 
In the query-agnostic map, we construct the candidate set through video features and modeling adjacent relations of video segments, then measure the mutual-matching similarity between video candidates and query vector. 
In the query-conditioned map, we leverage multi-modal features to construct a query-conditioned candidate set, then use IoU regression to predict the score of each proposal.
Although the proposal assessment methods of the two maps are different, they are complementary to each other. 
The query-agnostic map focuses on learning cross-modal similarity, while the query-conditioned map is aimed at implementing fusion alignment between different modalities. 
Theoretically, both of them represent candidate-level alignment representations, and candidates with the same coordinates represent the same video segment. 
Therefore, it is reasonable to make a joint prediction based on the candidate results of the two maps, by strengthening the connection between candidates. 
In particular, we employ specific optimization methods for the candidates in the two maps.
For the query-agnostic map, we enhance the text-video matching by employing a metric learning approach to compare the semantic similarity between different video segment candidates and their corresponding text vectors.
For the query-conditioned map, we utilize IoU regression to obtain the confidence scores for each candidate.
This approach allows for a more comprehensive and rational evaluation of the video segment candidates, thereby enabling more precise joint predictions and suitability for fine-grained makeup segment localization. 

The main contributions of this paper are summarized as follows:
\begin{itemize}
\item We present a novel framework named \textit{Dual-Path Temporal Map Optimization Network (DPTMO)} for make-up temporal video grounding, which jointly optimizes 2D temporal maps from query-agnostic and query-conditioned features to conduct a more comprehensive and reasonable evaluation of the segment candidates.
\item We propose an intra-modal feature enhancement strategy and a novel 2D temporal map optimization strategy, which effectively refine the extracted features and improve the quality of candidates. 
\item We conduct experiments on the challenging fine-grained YouMakeup dataset, demonstrating our model excels at processing detailed and subtle semantic information. 
Ablation studies and qualitative visualizations also verify the effectiveness of each component of the proposed model. 
\end{itemize}

\section{Related Work}
\subsection{Benchmarks for Instructional Video Understanding}
According to the content of videos, existing research on instructional video understanding can be divided into two categories: general domains and specific domains. 
In the general domains, researcher constructed datasets\cite{alayrac2016unsupervised,tang2019coin} with diverse themes to improve the generalization ability of models. 
These datasets usually involve various daily life activities such as nursing, vehicles, housework, and gadgets. 
For example, Alayrac~\etal~\cite{alayrac2016unsupervised} proposed a dataset that contains complex interaction relationships between people and objects. The videos are captured in a diverse range of indoor and outdoor settings.  
Coin~\cite{tang2019coin} is a large-scale instructional dataset designed to address the limitations of existing datasets in terms of diversity and size. It covers 180 different tasks in 12 domains related to daily life activities. 

For specific domains, researcher proposed datasets cover the activities in specific domains, such as cooking~\cite{regneri2013grounding,zhou2018towards}, sports narrative~\cite{yu2018fine}, furniture assembly~\cite{toyer2017human} and makeup~\cite{wang2019youmakeup}. 
Among these datasets, TACoS~\cite{regneri2013grounding} and YouCook2~\cite{zhou2018towards} datasets focus on the cooking domain. 
TACoS is a dataset consisting 127 videos depicting basic cooking activities, with an average length of 4.5 minutes. 
In contrast, YouCook2~\cite{zhou2018towards} is a much larger dataset that contains 2,000 videos with 176 hours. 
To sum up, these cooking datasets generally provide rich semantics of various culinary actions, it is worth noting that many of these actions are easily distinguishable from one another. 
Fine-grained Sports Narrative~\cite{yu2018fine} is a fined-grained video caption dataset in the sports narrative domain, focusing on the detailed motion of subjects and containing a detailed sports description. 
Furniture Assembly~\cite{toyer2017human} is a video dataset about the action of human assembling furniture. 
YouMakeup~\cite{wang2019youmakeup} is a large-scale dataset in the makeup domain with fine-grained temporal and spatial annotation. 
Different from other datasets overlooking closely-resembled information, videos in the YouMakeup dataset are captured with the same facial background and could only be distinguished by subtle differences in the instructor's expressions or actions. 
As a result, the model is required with a high-level, fine-grained semantic comprehension for both temporal and spatial contexts. 

\subsection{Video Grounding Methods}
Previous methods for video grounding can be categorized into two main categories, \ie, proposal-free methods and proposal-based methods. 
Proposal-free methods~\cite{yuan2019find,he2019read,wang2019language,zeng2020dense,chen2020rethinking,mun2020local} aim to build a proposal-free architecture that directly predicts the boundaries of the target moment from the multi-modal features. 
However, a critical challenge of proposal-free methods is to understand video content and appropriately align the semantics between cross-modal features. 
The most popular way is to adopt various attention mechanism~\cite{vaswani2017attention} to align the video and language. 
Early works used diverse ways to operate video attention on language features. 
Yuan~\etal~\cite{yuan2019find} proposed a co-attention-based framework which learned informative features from language and global video segment. 
Mun~\etal~\cite{mun2020local} presented a local-global video-text interaction network that utilized a sequential query attention module to model implicit linguistic information. 
While proposal-free methods were effective and flexible in dealing videos with various lengths, they tended to ignore the rich information between boundary moments, which was essential to understand fine-grained semantic information. 

Different from proposal-free methods, proposal-based methods~\cite{anne2017localizing,zhang2020learning,wang2022negative,xu2019multilevel} aimed to generate various video segment proposals, and ranked them based on the similarity between the query and generated proposals. 
Early works~\cite{anne2017localizing,gao2017tall} applied sliding windows to generate video segment proposals. 
However, sliding windows could only attach attention to the moments within its receptive field, and failed to observe the long-range temporal dependencies across multiple windows. 
Moreover, the proposals generated by sliding window are usually overlapped, which led to redundant calculations. 
To tackle these issues, anchor-based methods~\cite{chen2018temporally,yuan2019semantic} generated proposals by multi-scale anchors and maintained the sequential or hierarchical relationship between proposals. 
From a temporal perspective, TGN~\cite{chen2018temporally} utilized a fine-grained frame-by-word interaction and produced proposals by pre-setting anchors at each time step.
From a hierarchical perspective, SCDM~\cite{yuan2019semantic} employed a hierarchical temporal convolutional network, which generated multi-scale proposal candidates in different levels of the convolution module. 
In summary, most of proposal-based methods analyzed the video clip independently but ignored their adjacent relations between boundary moments, leading to the limited quality of proposals. 

To tackle these issues, Zhang~\etal~\cite{zhang2020learning} presented a 2D temporal map diagram (2D-TAN) to construct a candidate set, aimed at modeling the adjacent temporal relation between candidates. 
The coordinates of the 2D map represented the start and end index of segment candidates. 
The 2D temporal map can enumerate all possible candidates in arbitrary lengths, which enabled the model to capture more comprehensive context information.
Inspired by this work, MMN~\cite{wang2022negative} employed video features to generate candidates and utilized mutual-matching learning to rank candidates. 

\begin{figure*}[ht]
\centering
\includegraphics[width=1.0\linewidth]{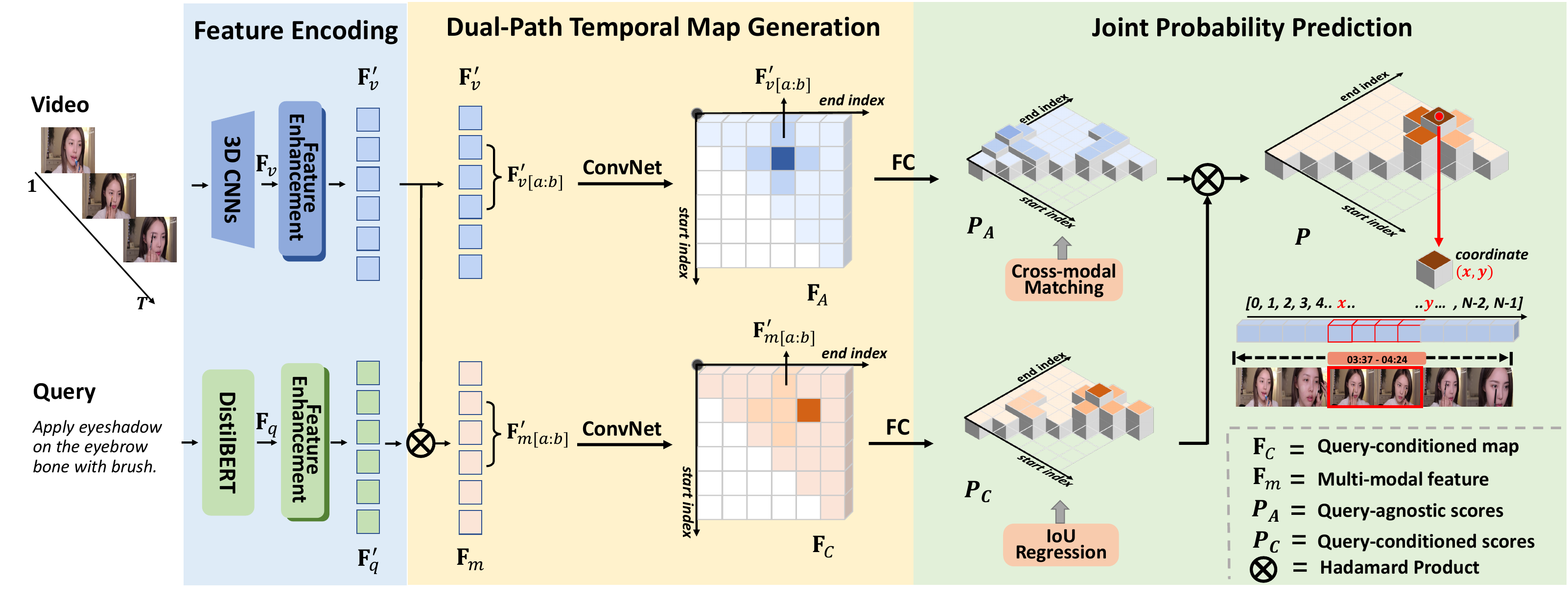}
\caption{The architecture overview of the proposed Dual-Path Temporal Map Optimization Network. In feature encoding, we first extract video and query features of the given video $V$ and query $Q$. Subsequently, query-agnostic map ${\bf F}_{A}$ and query-conditioned map ${\bf F}_{C}$ are generated in the dual-path temporal map generation module. Finally, we jointly leverage the query-agnostic and query-conditioned maps for proposal optimization.}
\label{fig:method}
\end{figure*}

As mentioned above, some progress has been made in the field of video grounding. 
However, MTVG is still in its infancy. 
It was proposed by~\cite{wang2019youmakeup} in the 4th PIC Challenge$\footnote{\url{https://github.com/AIM3-RUC/Youmakeup\_Challenge2022}}$. 
Given an untrimmed makeup video and a query describing makeup action, MTVG aims to localize the target makeup step in the video. 
This task requires the model to align fine-grained video-text semantics and distinguish makeup steps with subtle differences. 
MTVG is more challenging than video grounding due to its requirement for a higher level of fine-grained semantic understanding. 
Moreover, the length of the makeup video is longer than that of video grounding, which increases the difficulty of providing accurate long-term predictions. 
The MMN~\cite{wang2022negative} model was selected as the baseline for MTVG. 
Based on the 2D-TAN model~\cite{zhang2020learning}, MMN is inspired by metric learning method and leveraged negative inter-video visual-language pairs for the mutual matching. 
These negative samples can enhance the joint representation learning of two modalities via cross-modal mutual matching to maximize their mutual information. 
Although MMN achieved highly competitive performance in video grounding, there is still plenty of room for improvement in MTVG. 

\section{Proposed Model}
\subsection{Problem Formulation}\label{sec:prob}
In make-up videos, different instructional steps often share the same background, and the variations between them are often subtle and difficult to distinguish. 
The only way to distinguish them is pay attention to the makeup tools used or the resulting effect, such as the color of the eyelids after application. 
In this paper, we formulate MTVG as a dual-path temporal map optimization problem. 
Firstly, we extract video and query features, and then construct two types of 2D temporal maps, \ie, query-agnostic and query-conditioned temporal maps. 
The query-agnostic temporal map is built by visual features while query-conditioned temporal map is constructed by the multi-modal features. 
Each element in the temporal map represents a specific video segment. 
We then obtain the prediction of each candidate by fusing the 2D alignment score prediction maps. 
Finally, we deem the jointly optimized temporal map as the final result. 

\subsection{Feature Encoding}\label{sec:encoder}
\subsubsection{Language Encoder}
As shown in Fig.\ref{fig:method}, we utilize the DistilBERT~\cite{sanh2019DistilBERT} as the language encoder due to its lightweight design and excellent performance on NLP related tasks. 
Given a language query $Q$, we first convert it into a world-level token sequence and append a specific token `[CLS]' token at the beginning of the sequence. 
Then, the token sequence is fed into the DistilBERT model to extract the query feature sequence ${\bf F}_q\!=\!\{q_i\}_{i=1}^{L+1}$, where $q_i\in \mathbb{R}^{768}$, and $L$ denotes the length of token sequence. 
Considering the complexity of queries in MTVG, we feed the query feature sequence ${\bf F}_q$ into $H$ transformer encoder layers to capture sentence-level semantics. 
To obtain a representation of the whole sentence semantics, we utilize global average pooling on the enhanced query features to create the sentence-level query feature ${\bf F}_q^{\prime}\in \mathbb{R}^{d}$.

\subsubsection{Video Encoder}
Given a make-up video $V$, we first divide the input video into a clip sequence $V=\{v_{t}\}_{t=1}^{T}$, where $T$ is total clips and $v_{t}$ is the $t$-th clip. We use the pre-trained convolutional neural networks (I3D~\cite{carreira2017quo} or C3D~\cite{tran2015learning}) to extract clip-level visual feature. Therefore, we can get the visual feature ${\bf F}_v\!\in\!\mathbb{R}^{T\times d_v} $ with sequence length of $T$. 
To reduce the computation cost, we use a fully-connected layer to reduce the dimension of visual feature from $d_v$ to $d$. 
We use transformer encoder layers to enhance the representability of visual features. 
Firstly, we pass the extracted video feature ${\bf F}_{v}$ through $H$ Transformer encoder layers.
Subsequently, we use fixed-length sampling to control the number of video clips by setting the stride to $\frac{T}{N}$.  
Finally, we can obtain the enhanced video feature ${\bf F}_v^{\prime}\in \mathbb{R}^{N\times d}$,  ${\bf F}_v^{\prime} = \{v_i\}_{i=1}^{N}, v_i \in \mathbb{R}^d$. 

\subsection{Dual-Path Temporal Map Generation} 
Based on the above feature encoding, we obtain the enhanced video features ${\bf F}_v^{\prime}$ and query features ${\bf F}_q^{\prime}$. 
As mentioned in introduction, we use two types of temporal maps to generate proposals. 
The query-agnostic temporal map aims to select proposals through a metric-learning manner, while query-conditioned temporal map aims to select proposals by a regression manner. 
Both of these maps are chosen to be represented in the form of two-dimensional temporal maps in order to enumerate video segment candidates comprehensively. 
Although the features used to construct these two maps differ, the aggregated temporal map size remains consistent. 
Furthermore, elements at the same position represent the same video segment, enabling the feasibility of joint predictions.

\subsubsection{Query-agnostic Temporal Map}
\begin{figure}[t]
\centering
\includegraphics[width=1.0\linewidth]{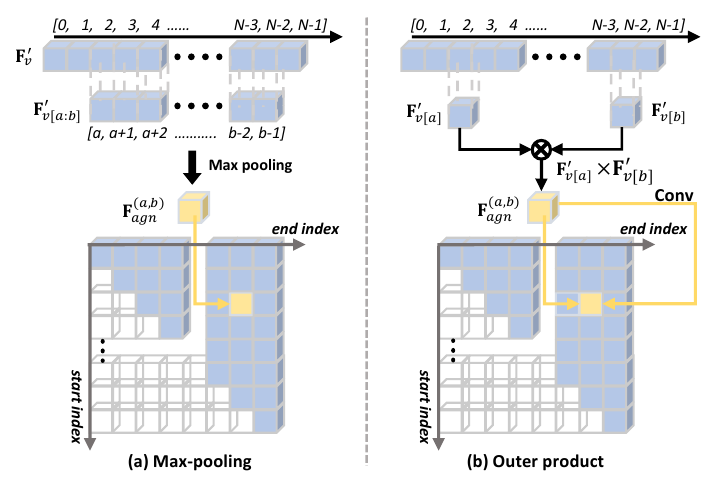}
\caption{The illustration of different 2D temporal map construction strategies: max-pooling (general) and outer product (ours).}
\label{fig:2D_map_aggregation_strategy}
\end{figure}

As shown in Fig.~\ref{fig:2D_map_aggregation_strategy} (a), if the candidate is with the span of $a$ to $b$, we can perform general operation (max-pooling) on the corresponding location in video features ${\bf F}_v^{\prime}$. 
The feature of candidates can be obtained by ${\bf F}_{A}^{(a,b)} = maxpooling({{\bf F}_v^{\prime}}_{[a]},{{\bf F}_v^{\prime}}_{[a+1]}, \ldots ,{{\bf F}_v^{\prime}}_{[b]})$, where $a$ and $b$ denote the indexes of start and end video clips, and $0\!\leq a\!\leq b \!\leq N\!-\!1$. 
However, we do not use the max pooling strategy to generate temporal map.

Current methods (\eg, \cite{zhang2020learning,wang2022negative}) often rely on max pooling to generate candidates and extract global semantic information from video clips due to its fast speed and small parameter size, which has limitations. 
Max pooling inevitably loses some information about different makeup steps during the calculation process, making it difficult for the model to distinguish the overall semantics of different video clips. 
The high-similarity candidates generated by max pooling affect the model’s ability to understand fine-grained semantics. 
Unlike max pooling, the outer product strategy focuses on \textit{the information at the video boundary moments}. 
This strategy increases the differences between generated candidates, making them easy to distinguish, and enables the model to capture key information of the video clips, thereby improving its ability to understand fine-grained semantics. Thus, we adopt the outer product strategy to generate temporal map.

As shown in Fig.~\ref{fig:2D_map_aggregation_strategy} (b), we create the 2D temporal map by ${\bf F}_{v}^{\prime} \times {{\bf F}_{v}^{\prime}}^\intercal $. 
Each element in the temporal map denotes a specific segment in all proposals, and we calculate it as: ${\bf F}_{A}^{(a,b)}\!=\!{{\bf F}_{v}^{\prime}}_{[a]} \times {{\bf F}_{v}^{\prime}}_{[b]}$.
The query-agnostic temporal map is denoted as ${\bf F}_{A} \in \mathbb{R}^{N\times N\times d}$, where $d$ indicates the feature dimension. 
We adopt $L_{A}$ layers of 2D convolution with the kernel size $K_{A}\times K_{A}$ to model the relations of proposal features ${\bf F}_{A} \in \mathbb{R}^{N\times N\times d}$. 
Zero padding is adopted in convolution to keep the output with the same shape with as ${\bf F}_{A}$, 
and this enables our model to keep a larger and stable receptive field for capture informative video features. 
Given the convolution structure, our model is capable to gradually model the correlation and learn the discrepancy between adjacent candidates. 

\subsubsection{Query-conditioned Temporal Map}
In query-agnostic temporal map, only video features are used to construct 2D temporal map, without considering the query information. 
We argue that query-agnostic temporal map is not sensitive to the query semantic information, which is hard to cope with fine-grained semantics. 
To this end, we built a query-conditioned temporal map based on \textit{the cross-modal features} of visual and textual modalities. 
In other words, this map is constructed by multi-modal clip features, which is the combination of video clip features and query representation.
We adopt Hadamard product to realize multi-modal fusion, the multi-modal feature ${\bf F}_{m}\!=\!{\bf F}_v^{\prime} \cdot {\bf F}_q^{\prime}$. 
Same as our query-agnostic temporal map mentioned above, query-conditioned temporal map adopts the outer product as its aggregation strategy, denoted as ${\bf F}_{C} \in \mathbb{R}^{N\times N\times d_{C}}$. 
We model these multi-modal relations between moments with $L_{C}$ layers of 2D convolution with kernel size $K_{C} \times K_{C}$, and adopt zero padding to keep its shape unchanged as well. 

\subsection{Joint Prediction Module}
We make a joint prediction, after obtaining the query-agnostic and query-conditioned temporal maps.
Firstly, we need to calculate the confidence score for video segment candidates in each maps.
For the query-agnostic temporal map, we use cosine similarity to estimate the matching quality of each proposals and queries. 
For the query-conditioned temporal map, we utilize fully-connected layers to reduce the dimension of candidate features to 1, representing their prediction scores.

In query-agnostic temporal map, there are two supervision signals to measure predictions and target moments (IoU regression, cross-modal matching), and we use two independent branches to map features to the same dimension. 
We adopt a linear layer to map the language features ${\bf F}_{q}^{\prime}$ into the same feature space $d_{H}$ and adopt the $1 \times 1$ convolution for the 2D video temporal map ${\bf F}_{A} \in \mathbb{R}^{N\times N\times d}$. 
We denote the scores of IoU regression and cross-modal matching as $s_{iou}$ and $s_{mm}$, respectively. 
The process is formulated as follows: 
\begin{equation}%
{{\bf F}_{q}^{iou}} =  W_{iou}{\bf F}_{q}^{\prime} + b_{iou}, 
{{\bf F}_{q}^{mm}} =  W_{mm}{\bf F}_{q}^{\prime} + b_{mm}, 
\end{equation}
\begin{equation}
\mathbf{\bf F}_{A}^{iou} = conv_{iou}({\bf F}_{v}^{\prime}), 
\mathbf{\bf F}_{A}^{mm} = conv_{mm}({\bf F}_{v}^{\prime}), 
\end{equation}
\begin{equation}
S_{iou} = {\mathbf{\bf F}_{A}^{iou}}^\intercal {{\bf F}_{q}^{iou}},
\end{equation}
\begin{equation}%
S_{mm} = {\mathbf{\bf F}_{A}^{mm}}^\intercal {{\bf F}_{q}^{mm}},
\end{equation} 
where the stride and kernel size of convolution layers $conv_{iou}$ and $conv_{mm}$ are set to 1. $W_{iou}$, $W_{mm}$ and $b$ are learnable parameters. The 2D temporal maps ${\bf F}_{A}^{iou}, {\bf F}_{A}^{mm}\in \mathbb{R}^{N\times N\times d_{H}}$.

Since the range of cosine similarity score is $s_{iou}\in S_{iou}$ in the range of $(-1,1)$, the supervision signal (IoU signal) is in the range of $(0,1)$. We adopt the sigmoid function $\sigma$ to change the value distribution to make it near the suitable neutral region. 
To let the prediction score $s_{iou}$ occupy most of the region in $(0, 1)$, we amplify the $s_{iou}$ by a factor of 10.
Similarly, to accommodate the distribution of IoU, we use linear transformation to map $s_{mm} \in S_{mm}$ from $(-1,1)$ to $(0,1)$, and adopt exponentiation to make the distribution more reasonable, where $u$ represents the power. 
The prediction of a candidate in query-agnostic temporal map is denoted as $p_{A}$. $p_{mm}$ and $p_{iou}$ are the matching scores based on cross-modal mutual matching and IoU regression, respectively.  
$u$ is the power of cross-modal matching scores, which is set to 0.3. 
\begin{equation}%
p_{mm} = {(s_{mm} \times 0.5 + 0.5)}^{u},
\end{equation}
\begin{equation}%
p_{iou} = \sigma(10 \cdot s_{iou}),
\end{equation}
\begin{equation}%
p_{A} = p_{mm} \times p_{iou}.
\end{equation}

For query-conditioned temporal map, we feed ${\bf F}_{C} \in \mathbb{R}^{N\times N\times d_{C}}$ to a fully-connected layer with a sigmoid function $\sigma$ to generate a 2D score map. 
$P_{C}^{\prime}$ represents the final matching score for each candidates in query-conditioned temporal map: 
\begin{equation}
\mathbf{S}_{C} =  FC({\bf F}_{C}),
\end{equation}
\begin{equation}
P_{C}^{\prime} = \sigma(10 \cdot {S}_{C}), 
\end{equation}
where $FC$ denotes the fully-connected layer which transforms the query-conditioned feature channel dimension from $d_{C}$ to $1$ and get $\mathbf{S}_{C}\!\in\!\mathbb{R}^{ N\times N \times 1}$. 

2D temporal map can enumerate all possible proposals with various lengths, yet bringing a concern of large computational cost. 
We follow 2D-TAN~\cite{zhang2020learning} to adopt the sparse sampling strategy to reduce redundant candidate moments. 
Specifically, we first enumerate all possible candidates with short duration, then gradually increase the time interval to the length of the candidate's moments. 

All the valid scores in the map are collected, the query-agnostic and query-conditioned scores are denoted as $P_{A} = {\{ p_{A}^{i}\}}_{i=1}^{M}$ and $P_{C} = {\{ p_{C}^{i}\}}_{i=1}^{M}$, where $M$ is total number of valid candidates. 
We adopt the product of above two map scores as the final prediction, denote as $p^{i} = p_{A}^{i} \times p_{C}^{i}$. The set of all candidates prediction is denoted as $P = {\{ p^{i} \}}_{i=1}^{M}$. 
During the inference, we apply the non maximum suppression with a threshold $nms$ to remove redundant proposal candidates.

\begin{figure}[t]
\centering
\includegraphics[width=1.0\linewidth]{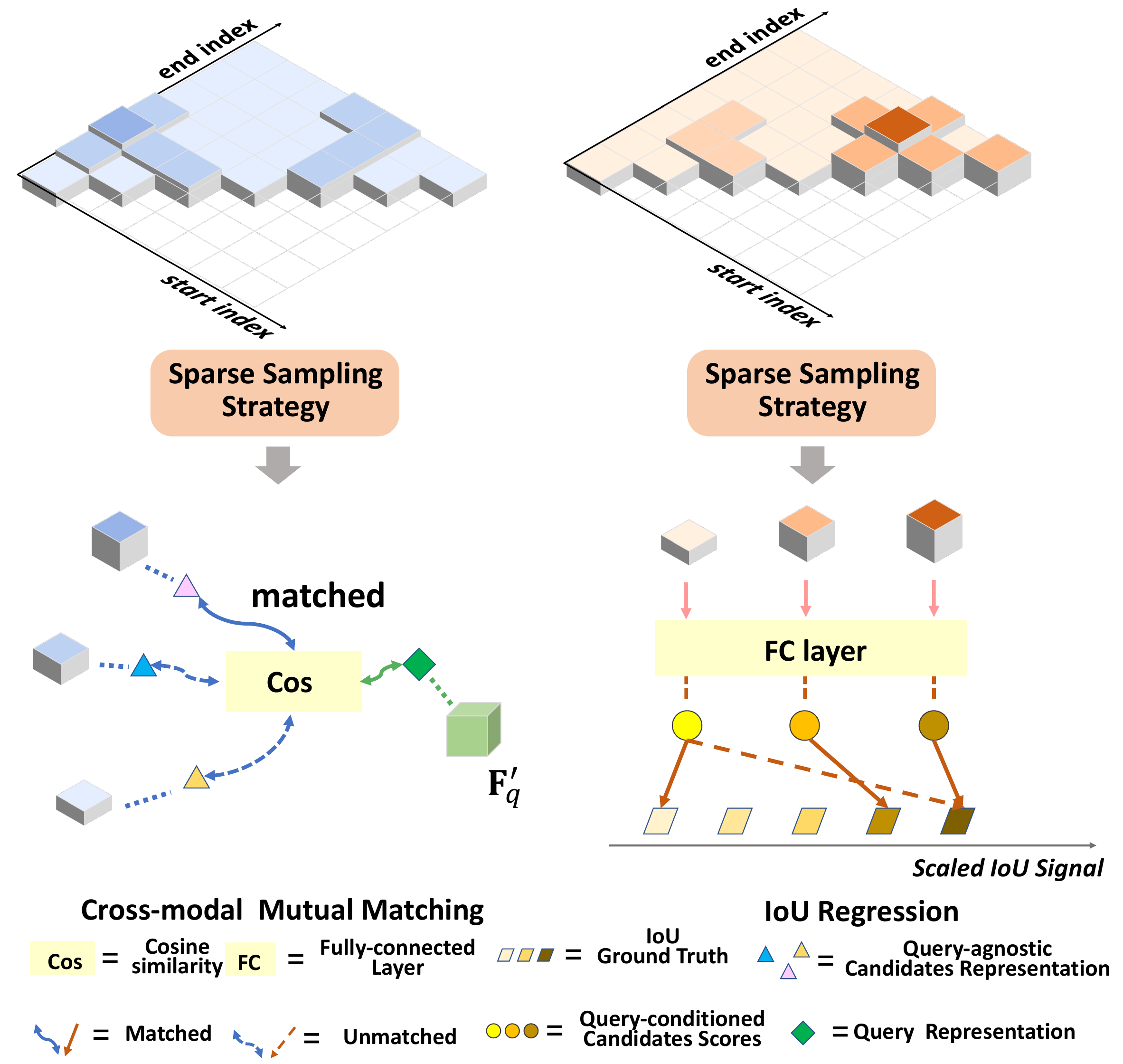}
\caption{The illustration of different loss optimizations: Cross-modal mutual matching (left) and IoU regression (right).}
\label{fig:loss_illu}
\end{figure}

\subsection{Loss Optimization}
To optimize the proposed model, we adopt a dual-path loss $\mathcal{L}$ including a multi-task loss $\mathcal{L}_{A}$ for the query-agnostic temporal map and a multi-modal loss $\mathcal{L}_{C}$ for query-conditioned map.  
The dual-path loss $\mathcal{L}$ is formulated as follows:
\begin{equation}%
\mathcal{L} = \mathcal{L}_{A}+\mathcal{L}_{C}.
\end{equation}

\subsubsection{Loss for Query-agnostic Temporal Map}
Following MMN~\cite{wang2022negative}, we adopt two kinds of supervision signals (IoU regression and cross-modal matching) for query-agnostic temporal map: 
\begin{equation}%
\mathcal{L}_{A} = \mathcal{L}_{iou}+\lambda \mathcal{L}_{mm}.
\label{eq:loss_a}
\end{equation}

\textbf{IoU Regression} 
As shown in Fig.~\ref{fig:loss_illu}, we employ the scaled IoU values $y_{i}$ as the supervision signal. 
IoU $o_{i}$ is calculated based on the temporal information of video segment represented by candidates and the ground truth video segment. IoU reflects the degree of overlap between them. 
We linearly scale these IoU values $y_{i}$ from $(t_{min}, t_{max})$ to $(0,1)$, where $t_{min}$ and $t_{max}$ are the thresholds set by ourselves. The detail method to scale the IoU scores $y_{i}$ is as follows:
\begin{equation}
y_{i}=\left\{
\begin{aligned}
&\ 0  &\  o_{i}<=t_{min},\\
&\ \frac{o_{i}-t_{min}}{t_{max}-t_{min}} &\   t_{min}<o_{i}<=t_{max},\\
&\ 1 &\  o_{i}>=t_{max}.\\
\end{aligned}
\right.
\label{eq:IoU_scaled}
\end{equation}

We use the binary cross entropy loss to constrain the IoU regression. The loss function is formulated as follows: 
\begin{equation}%
\mathcal{L}_{iou} = -\frac{1}{M} \sum\nolimits_{i=1}^{M}(y_{i}\log p_{iou}^i+(1-y_{i})\log(1-p_{iou}^i)), 
\end{equation}
where $M$ denotes the total number of valid candidates and $p_{iou}^i$ is the IoU score prediction of $i-$th proposal in the query-agnostic temporal map. 

\begin{table*}[t!]
\centering
\footnotesize
\caption{The comparison of our model (DPTMO) and the baseline (MMN) on the YouMakeup dataset.}
\setlength{\tabcolsep}{4.5mm}{
\begin{tabular}{ c|c|ccc|ccc}
\hline\hline
\multirow{2}{*}{Model}   & \multirow{2}{*}{Visual feature}  &  & R@1 &  &  & R@5 & \\ 
&  & IoU@0.3 $\uparrow$ & IoU@0.5$\uparrow$ & IoU@0.7$\uparrow$ & IoU@0.3$\uparrow$ & IoU@0.5$\uparrow$ & IoU@0.7$\uparrow$\\
\hline
LGI~\cite{mun2020local} & C3D & 30.60 & 16.45 & 6.19 & N/A & N/A & N/A \\ 
MMN~\cite{wang2022negative}      &  C3D  & 33.47 & 23.05 & 11.78 & 63.28 & 48.88 & 25.07    \\ 
\textbf{DPTMO (Ours)} &  C3D  & \textbf{45.53} & \textbf{33.63} & \textbf{19.23} & \textbf{73.51} & \textbf{60.66} & \textbf{33.91}\\ \hline
LGI~\cite{mun2020local} & I3D & 35.36 & 21.00 & 8.77 & N/A & N/A & N/A \\ 
MMN~\cite{wang2022negative}      &  I3D  & 48.09 & 35.18 & 20.08 & 76.79 & 64.13 & 36.25    \\
\textbf{DPTMO (Ours)} &  I3D  & \textbf{61.70} & \textbf{47.77} & \textbf{28.99} & \textbf{83.68} & \textbf{73.82} & \textbf{43.29} \\  
\hline\hline
\end{tabular}
}
\label{tab:comparision}
\end{table*}

\textbf{Cross-modal Mutual Matching} 
Following MMN, we adopt a metric learning approach to realize the cross-modal matching. 
In particular, we generate the moment-sentence pairs with their negative pairs as our supervision signals which are sampled from both \textit{inter} and \textit{intra} videos. 
Given a language query, the target video segment is deemed as the positive sample. 
The negative video moments are sampled from other low IoU moments inside the video (\textit{intra}) and moments from other videos (\textit{inter}). 
The negative sentences are sampled from other sentences in the video (\textit{intra}) and from other videos (\textit{inter}). 
The cross-modal pair discrimination loss~\cite{li2020learning} is adopted to judge semantic similarity between video features and sentence features. 
As shown in Fig.~\ref{fig:loss_illu}, we use the moment features ${\bf F}_{A}^{mm}$ and sentence features ${{\bf F}_{q}^{mm}}$ to operate cross-modal mutual matching in a non-parametric softmax form. 
We denote the two scalar product values (sentence features with the negative moment samples, moment features with the negative sentence samples) as $Pr_q = {{{\bf F}_{q}^{mm}}}_{i}^\intercal\mathbf{f}^v_{mm}$ and ${Pr_v = \mathbf{f}^v_{mm}}_i^\intercal {{{\bf F}_{q}^{mm}}}$ respectively. 
This process is formulated as follows: 
\begin{equation}%
p(i_{s}|v) = \frac{exp((Pr_q-m)/\tau_{v})}{exp((Pr_q-m)/\tau_{v}) + \sum_{j\neq i}^{N_{s}}exp(Pr_q/\tau_{v}) } 
\label{eq:loss_mutual_v}
\end{equation}
\begin{equation}%
p(i_{v}|s) = \frac{exp((Pr_v-m)/\tau_{s})}{exp((Pr_v-m)/\tau_{s}) + \sum_{j\neq i}^{N_{v}}exp(Pr_v/\tau_{s}) },
\label{eq:loss_mutual_s}
\end{equation} 
where the $i$-{th} moment or sentence represents a instance-level class $i_{v}$ or $i_{s}$, $\tau_{s}$ and $\tau_{v}$ denote the temperatures and $N_{s}$ and $N_{v}$ is the total numbers of the samples instances in the specific batch. 
$m$ represents the threshold for negative sample determination. Moments with an IoU lower than $m$ are considered negative samples.

Cross-modal mutual matching is aimed to maximize the likelihood $\prod^{N}_{i=1}p(i_s|v_i)\prod^{N}_{i=1}p(i_v|s_i)$ where $N$ represents the total number of moment-sentence pairs for training. 
The cross-correlation measures the similarity between different modalities features and makes our model to seize the mutual information with the help of the binary pair supervision. 
The loss function is formulated as follows:
\begin{equation}
\mathcal{L}_{mm} = -(\sum_{i=1}^{N}logp(i_{v}|s_{i})+\sum_{i=1}^{N}logp(i_{s}|v_{i})).
\end{equation} 

\subsubsection{Loss for Query-conditioned Temporal Map}\label{sec:signal}
We choose the scaled IoU value $y_{i}$ as the supervision signal in query-conditioned temporal map, which is the same as IoU regression in query-agnostic map.
We utilize a binary cross entropy loss to train the query-conditioned temporal map branch in our model. The loss function $\mathcal{L}_{C}$ is formulated as follows:
\begin{equation}%
\mathcal{L}_{C} = -\frac{1}{M} \sum_{i=1}^{M}(y_{i}\log p_{C}^i+(1-y_{i})\log(1-p_{C}^i)),
\end{equation} 
where $p_{C}^i$ is the prediction of $i-$th moment in query-conditioned map and $M$ is the number of valid candidates.

\section{Experiments}
\subsection{Experiment Setup} 
\subsubsection{Dataset} 
\textbf{YouMakeup}~\cite{wang2019youmakeup} is a large-scale instructional video dataset used to support the research of makeup action understanding. 
Makeup videos naturally contain rich fine-grained semantic information. 
For example, each step in the same makeup video is mainly distinguished by the different actions, the makeup tools, and the makeup regions. 
However, analyzing makeup videos can be challenging due to the tiny size of makeup regions and the shared facial background across different makeup actions. 
The YouMakeup dataset consists of 2,800 videos crawled from the YouTube website with 420 hours totally. 
These videos are sliced into different instructional steps and annotated with specific natural language sentences. 
There are 30,626 steps, with 10.9 steps on average for each video. 
The length of makeup video varies from 15 seconds to 1 hour and 9 minutes.
The significant varied duration and dense steps make the YouMakeup dataset more complex and bring more challenges for semantic understanding. 

\begin{table*}[t]
\centering
\caption{Ablation study of feature enhancement module in feature encoding on the YouMakeup dataset.}
\setlength{\tabcolsep}{4.25mm}{
\begin{tabular}{c|c|ccc|ccc}
\hline\hline
\multirow{2}{*}{Model}  & \multirow{2}{*}{Visual feature}   &   & R@1 &  &  & R@5 & \\ 
  &   & IoU@0.3$\uparrow$ & IoU@0.5$\uparrow$ & IoU@0.7$\uparrow$ & IoU@0.3$\uparrow$ & IoU@0.5$\uparrow$ & IoU@0.7$\uparrow$  \\ \hline
  DPTMO w/o intra-v & C3D & 42.34 & 31.17 & 17.15 & 71.01 & 57.97 & 30.57    \\ 
DPTMO w/o intra-q & C3D & 42.97 & 30.38 & 16.45 & 71.42 & 58.35 & 31.83   \\ 
\textbf{DPTMO (Ours)} & C3D & \textbf{45.53} & \textbf{33.63} & \textbf{19.23} & \textbf{73.51} & \textbf{60.66} & \textbf{33.91}\\   \hline
DPTMO w/o intra-v & I3D & 55.04 & 41.24 & 22.80 & 80.58 & 68.90 & 37.92 \\ 
DPTMO w/o intra-q & I3D & 58.10 & 42.44 & 25.04 & 81.47 & 69.97 & 40.83 \\ 
\textbf{DPTMO (Ours)} & I3D & \textbf{61.70} & \textbf{47.77} & \textbf{28.99} & \textbf{83.68} & \textbf{73.82} & \textbf{43.29}  \\ 
\hline\hline
\end{tabular}
}
\label{tab:feature_enhancement}
\end{table*}

\subsubsection{Evaluation Metrics}
Following previous work~\cite{anne2017localizing,gao2017tall}, we adopt the metric ``R@$N$, IoU@$\theta$'' to evaluate the performance of the proposed model. 
``R@$N$, IoU@$\theta$'' denotes the ratio of temporal Intersection over Union (IoU) with ground truth is larger than threshold $\theta$ in the top-$N$ predicted moments. 
The higher IoU means the proposed model better understands fine-grained semantic information and can distinguish the subtle difference. 
Following previous work~\cite{zhang2020learning,wang2022negative}, we set $N \in \{1, 5\}$, and $\theta \in \{0.3, 0.5, 0.7\}$

\subsubsection{Implementation Details}
For the query encoder, we use the HuggingFace implementation$\footnote{
\url{
https://huggingface.co/docs/transformers/model\_doc/distilbert}}$ of DistilBERT with the pre-trained model ``DistilBERT-base-uncased''. 
The number of transformer encoder layers $H$ in video and language encoders is set to $2$ in our experiments. 
The ablation study of $H$ is discussed in Sec.~\ref{sec:abl}. 
For the input video, we use the pre-trained 3D convolutional neural networks (I3D~\cite{carreira2017quo} or C3D~\cite{tran2015learning}) as the backbone to exact video features. 
For the proposed DPTMO model, we adopt an 4-layer convolution network with kernel size of 3 in query-agnostic temporal map (\ie, $L_{A}$=4, $K_{A}$=3), and an 3-layer convolution network with kernel size of 3 in query-agnostic temporal map (\ie, $L_{C}$=3, $K_{C}$=3). 
The hidden size $d$ and $d_{C}$ are set to 512 and 128, respectively. 
For query-agnostic temporal map, we set temperature to $\tau_{s}$=$\tau_{v}$=0.05 in Eqs.~\ref{eq:loss_mutual_v}-\ref{eq:loss_mutual_s}. 
We choose $0.5$ as the lower bound for negative moment sampling, and set the margin in the pair discrimination loss to $0.1$ for cross-modal matching. 
For the query-conditioned temporal map, the scaling thresholds $t_{min}$ and $t_{max}$ (in Eq.~\ref{eq:IoU_scaled}) are set to 0.3 and 0.7 in the IoU regression. 
For training details, the proposed model is optimized by the Adam optimizer~\cite{kingma2014adam} with the learning rate of $10^{-4}$, and the batch size of $8$. 
The weight of pair discrimination loss $\lambda$ in Eq.~\ref{eq:loss_a} is set to 0.01. 
The threshold $nms$ of Non Maximum Suppression (NMS) is set to 0.4 during the inference.

\begin{table*}[t]
\footnotesize
\centering
\caption{Ablation study of each component in our method with C3D and I3D features on the YouMakeup dataset.}
\setlength{\tabcolsep}{2.6mm}{
\begin{tabular}{ c|c c c|c c c|c c c}
\hline\hline
\multirow{2}{*}{Visual feature} & Outer   &  Dual-Path   & Feature &  & R@1 &  &  & R@5 & \\ 
           & Product & Optimization & Enhancement & IoU@0.3$\uparrow$ & IoU@0.5$\uparrow$ & IoU@0.7$\uparrow$ & IoU@0.3$\uparrow$ & IoU@0.5$\uparrow$ & IoU@0.7$\uparrow$\\
\hline
C3D  &   &   &              & 33.47 & 23.05 & 11.78 & 63.28 & 48.88 & 25.07 \\  
C3D  &  \checkmark &  &     & 35.71 & 24.57 & 12.72 & 65.08 & 49.19 & 24.38    \\   
C3D  & \checkmark  &  \checkmark  &   &  43.38 & 29.93 & 16.86 & 69.94 & 55.23 & 30.63 \\ 
C3D  &  \checkmark &  \checkmark &  \checkmark  & \textbf{45.53} & \textbf{33.63} & \textbf{19.23} & \textbf{73.51} & \textbf{60.66} & \textbf{33.91}  \\
\hline
I3D  &   &   &              & 48.09 & 35.18 & 20.08 & 76.79 & 64.13 & 36.25    \\  
I3D  &  \checkmark &  &     & 49.89 & 37.86 & 21.69 & 76.00 & 62.61 & 34.54     \\  
I3D  & \checkmark  & \checkmark  &    & 53.43 & 39.85 & 23.02 & 79.10 & 66.31	& 37.73   \\
I3D  &  \checkmark &  \checkmark & \checkmark & \textbf{61.70} & \textbf{47.77} & \textbf{28.99} & \textbf{83.68} & \textbf{73.82} & \textbf{43.29}  \\

\hline\hline
\end{tabular}
}
\label{tab:ablation_exponent}
\end{table*}

\subsection{Main Comparison}
To validate the effectiveness of our proposed method, we compare it with MMN~\cite{wang2022negative}, an existing state-of-the-art proposal-based method. 
As shown in Table~\ref{tab:comparision}, our approach achieves much better results on the YouMakeup dataset with both C3D and I3D features.  
For the C3D feature, our model exhibits large improvements on all metrics. Specifically, our model achieves ``R@1, IoU@0.7'' of $19.23\%$, and ``R@5, IoU@0.7'' of $33.91\%$. When compared with MMN, our DPTMO achieves improvements of $63.24\%$ and $35.26\%$ on ``R1@1, IoU@0.7'' and ``R@5, IoU@0.7'', respectively. 
We can observe that the performance of our model based on I3D features is improved by about ten percents for each metric. 
Our model obtains $61.70\%$ on ``R@1, IoU@0.3'', $47.77\%$ on ``R@1, IoU@0.5'', and $28.99\%$ on ``R@1, IoU@0.7''. 
This model surpasses the baseline, MMN, by a large margin (\ie, 44.37\% improvements on ``R@1, IoU@0.7'', 19.42\% improvements on ``R@1, IoU@0.7''). 

We also train and evaluate the performance of the advanced proposal-free method LGI~\cite{mun2020local} on the YouMakeup dataset, with careful fine-tuning. We can see that LGI only achieves 30.60 and 35.36 on ``R@1, IoU@0.3'' with the C3D and I3D features, respectively. This result reflects that the generic video grounding method LGI is inferior to our method for the fine-grained makeup video grounding task. 
In summary, these comparisons quantitatively demonstrate that our model is capable of leveraging the benefits of dual-path temporal map optimization, significantly improving the performance of makeup video analysis. 

\subsection{Ablation Studies}\label{sec:abl}
In this section, we present ablation studies on the YouMakeup dataset to evaluate the effectiveness of the proposed model. 
Specifically, we conduct ablation studies from the following aspects: 1) feature enhancement, 2) each component of the proposed model, and 3) hyper-parameter $H$ of transformer encoder. 
\subsubsection{Feature Enhancement}  
As discussed in Sec.~\ref{sec:encoder}, the transformer encoder layer is utilized to enhance the feature representability of each modality. 
To evaluate the effectiveness of transformer encoder layers, we conducted experiments on the YouMakeup dataset using both I3D and C3D features. 
In particular, we examined two variants of the proposed DPTMO: ``DPTMO w/o intra-v'' removes the feature enhancement on the visual modality, and ``DPTMO w/o intra-q'' removes the feature enhancement on the textual modality. 
As shown in Table~\ref{tab:feature_enhancement}, for the I3D feature, the worst performance is achieved by ``DPTMO w/o intra-v", and the performance is dropped by 22.80\% at ``R@1, IoU@0.7'' and 37.92\% at ``R@5, IoU@0.7''. 
For the C3D feature, the worst performance is also achieved by ``DPTMO w/o intra-v'', and the performance is dropped by 42.34\% at ``R@1, IoU@0.3'' and 30.57\% at ``R@5, IoU@0.7''. 
This results demonstrate that feature enhancement is beneficial in  enhancing the contextual relationships between adjacent features.  
Additionally, when compared with ``DPTMO w/o intra-q'' model, the performance of ``DPTMO w/o intra-v'' in C3D and I3D features is significant reduced, indicating that the visual feature is more crucial than the textual feature for MTVG.

\subsubsection{Each Component of the Proposed Model}
We evaluate the effectiveness of each component in the proposed model from three aspects: 1) 2D temporal map construction strategy, 2) dual-path optimization, and 3) feature enhancement. 
The experiments are conducted on the YouMakeUp dataset with both C3D and I3D features. 
As shown in Table~\ref{tab:ablation_exponent}, row 1 and row 5 list the baseline results of our model, which adopt max-pooling as the 2D map construction strategy. 
As shown in row 2 and row 6, the temporal generation strategy in baseline is replaced with outer product. We can see that the performance is improved, \eg, ``R@1, IoU@0.7'' is improved from 20.08\% to 21.69\% in terms of I3D feature, and from 11.78\% to 12.72\% in terms of C3D feature. 
These results indicate that the outer product increases the discrepancy between candidates, making them more distinct and refining their quality, thus improve the model's ability to evaluate candidates. 

Row 3 and row 7 report the model with outer product and dual-path temporal optimization. 
We can observe that the model performs 53.43\%, 39.85\% and 23.02\% in ``R@1, IoU@0.3'', ``R@1, IoU@0.5'', ``R@1, IoU@0.7'' respectively in I3D feature. 
This result means that compared to the single path model method, the dual path structure model has more stable performance and more accurate prediction.
We will explain the specific reasons in Sec.~\ref{sec:visualize_dual_path}. 

Finally, row 4 and row 8 present the model with feature enhancement using the transformer encoder layer. 
The proposed method achieves the best performances with the ``R@1, IoU@0.3'' of 61.70\%, ``R@1, IoU@0.5'' of 47.77\%, ``R@1, IoU@0.7'' of 28.99\% based on I3D feature. 
These results demonstrate that feature enhancement leads to a significant improvement in the contextual feature of proposals, thereby enhancing the overall performance of the model. 

\begin{table}[t]
\centering
\caption{Ablation study of the number of transformer layers with C3D and I3D features.}
\resizebox{1.0\linewidth}{!}{
\begin{tabular}{c|ccc|ccc}
\hline\hline
 &   & R@1 &  &  & R@5 & \\ 
    & IoU@0.3$\uparrow$ & IoU@0.5$\uparrow$ & IoU@0.7$\uparrow$ & IoU@0.3$\uparrow$ & IoU@0.5$\uparrow$ & IoU@0.7$\uparrow$  \\ \hline
 \multicolumn{7}{c}{C3D features}   \\ \hline
  $H$=0 & 43.38 & 29.93 & 16.86 & 69.94 & 55.23 & 30.63  \\ 
 $H$=1 & \textbf{45.72} & 32.49 & 17.68 & 72.56 & 59.49 & 31.29  \\ 
 \textbf{$H$=2} & 45.53 & \textbf{33.63} & \textbf{19.23} & \textbf{73.51} & \textbf{60.66} & \textbf{33.91}  \\ 
 $H$=3 & 43.20 & 31.61 & 16.45 & 70.38 & 58.83 & 31.95  \\ 
 $H$=4 & 44.55 & 30.94 & 17.05 & 72.25 & 57.12 & 31.51  \\ \hline
\multicolumn{7}{c}{I3D features}   \\ \hline
 $H$=0 & 53.43 & 39.85 & 23.02 & 79.10 & 66.31 & 37.73 \\
 $H$=1 & 58.83 & 45.91 & 27.47 & 82.41 & 71.27 & 41.81  \\ 
 \textbf{$H$=2} & \textbf{61.70} & \textbf{47.77} & 28.99 & 83.68 & \textbf{73.82} & 43.29  \\ 
 $H$=3 & 60.31 & 45.82 & 28.13 & \textbf{83.83} & 72.47 & 43.04  \\ 
 $H$=4 & 59.84 & 46.98 & \textbf{29.40} & 83.23 & 73.63 & \textbf{44.04} \\
\hline\hline
\end{tabular}
}
\label{tab:I3D_num_trans_layers}
\end{table}

\subsubsection{Hyper-parameter $H$ in Transformer Encoder} 
Here, we discuss the influence of hyper-parameter $H$ in transformer encoders. 
As shown in Table~\ref{tab:I3D_num_trans_layers}, we report the experimental results with C3D and I3D features under different settings (\ie, $H\in \{0, 1, 3, 3, 4\}$). 
We can conclude that using more transformer encoder layers can effectively enhance the intra-modal relationship within each modality. However, too many encoder layers may lead to overfitting of the model, which results in performance degradation. 
Therefore, we set $H=2$ as the optimal setting. 

\subsection{Qualitative Visualization and Analysis}
\subsubsection{Visualization of Different 2D Temporal Map Construction Strategies}
\begin{figure}[t!]
\centering
\includegraphics[width=1.0\linewidth]{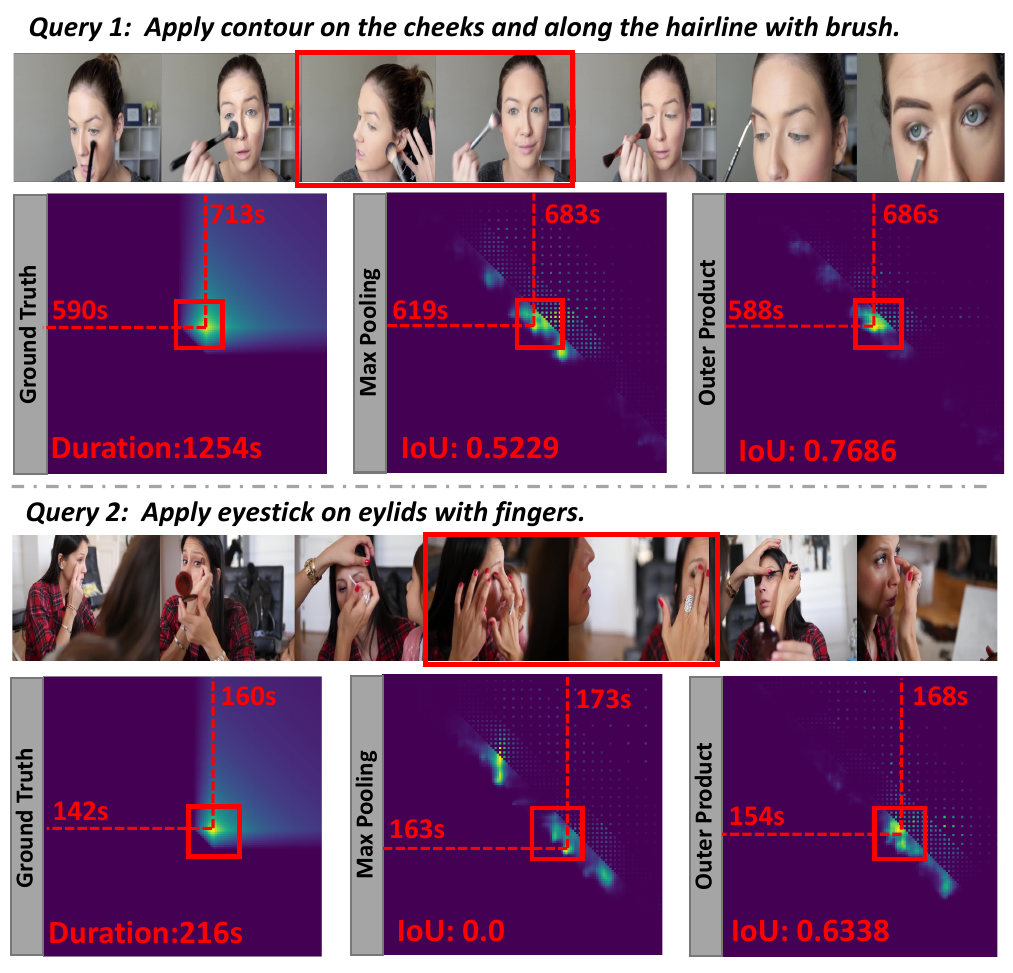}
\caption{Qualitative comparison of 2D maps between max pooling (general) and outer product (Ours). IoU denotes the temporal IoU value between the predicted top-1 result and ground truth.}
\label{fig:map_cmp_2Dmap_construction}
\end{figure}

Here, we give some visualization results to verify the effectiveness of the 2D map construction strategy in our model. We test the baseline MMN using either the outer product or max pooling aggregation methods.
As shown in Fig.~\ref{fig:map_cmp_2Dmap_construction}, each row represents one example, we visualize 2D temporal map scores generated by the max-pooling or outer product strategy. 
Compared with max-pooling strategy, the outer product produces less noise candidates and allocates more attention to the correct candidates. 
In $Query$ 1, we can observe that the baseline MMN with outer product cab localize the correct video segment more accurately and has a higher IoU.
It indicates that outer product strategy is more sensitive to the fine-grained information and capable to improve the quality of candidates. 
Additionally, there are some interesting results. 
As shown in $Query$ 2, if the model pays excessive attention to the wrong candidate moments, it will localize a completely wrong segment and the IoU drops to 0. This is our motivation to use outer product and dual path optimization to reduce noise.  

\begin{figure*}[t!]
\centering
\includegraphics[width=1.0\linewidth]{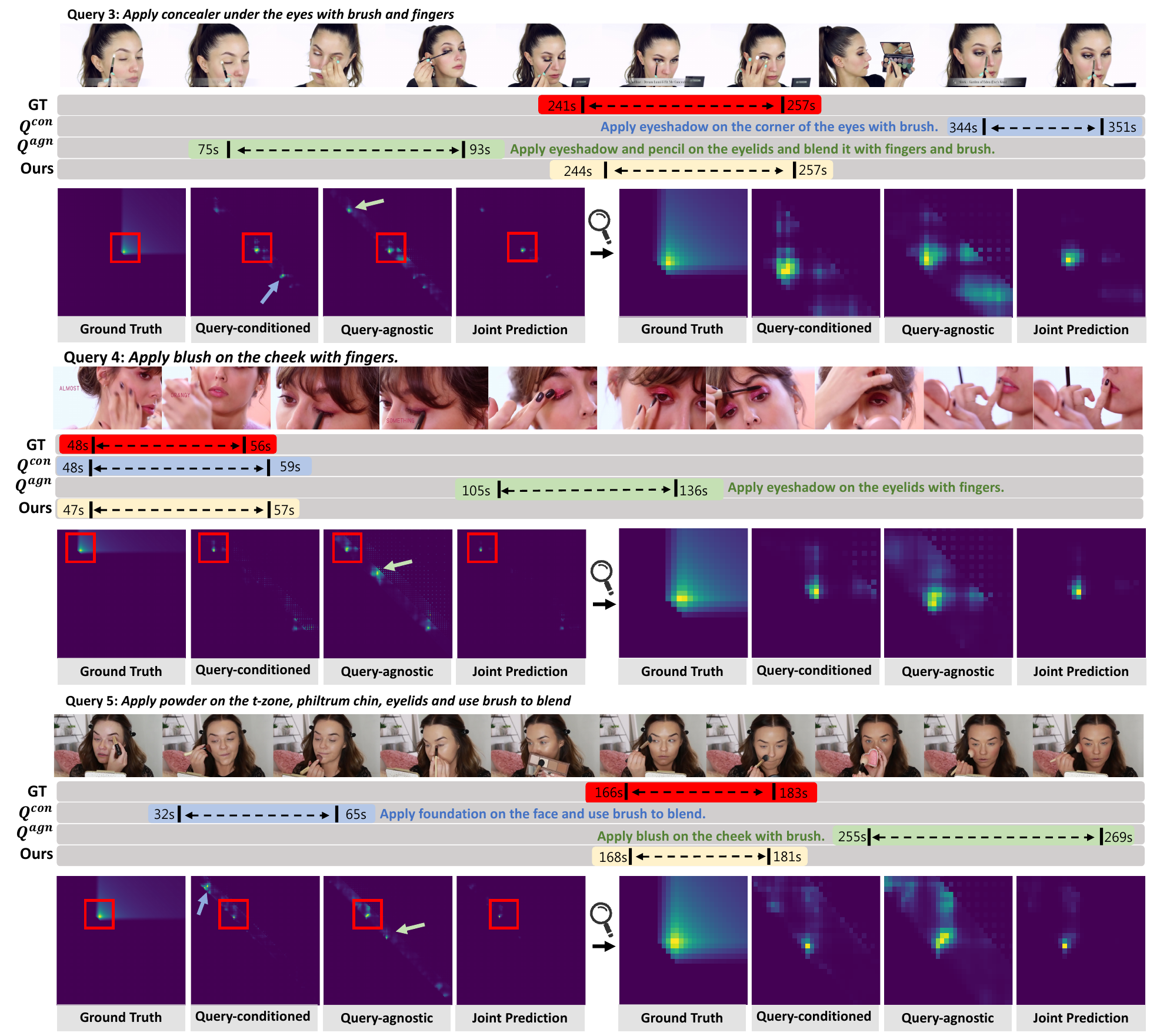}
\caption{Qualitative results of predicted 2D temporal maps. 
GT denotes the ground truth temporal map, $Q^{agn}$ and $Q^{con}$ denote the interference proposal in the query-agnostic and query-conditioned temporal maps, respectively. Ours denotes the prediction results of our proposed method. 
We can see that our method enjoys better grounding results. }
\label{fig:map_cmp}
\end{figure*}

\begin{figure*}[t!]
\centering
\includegraphics[width=1.0\linewidth]{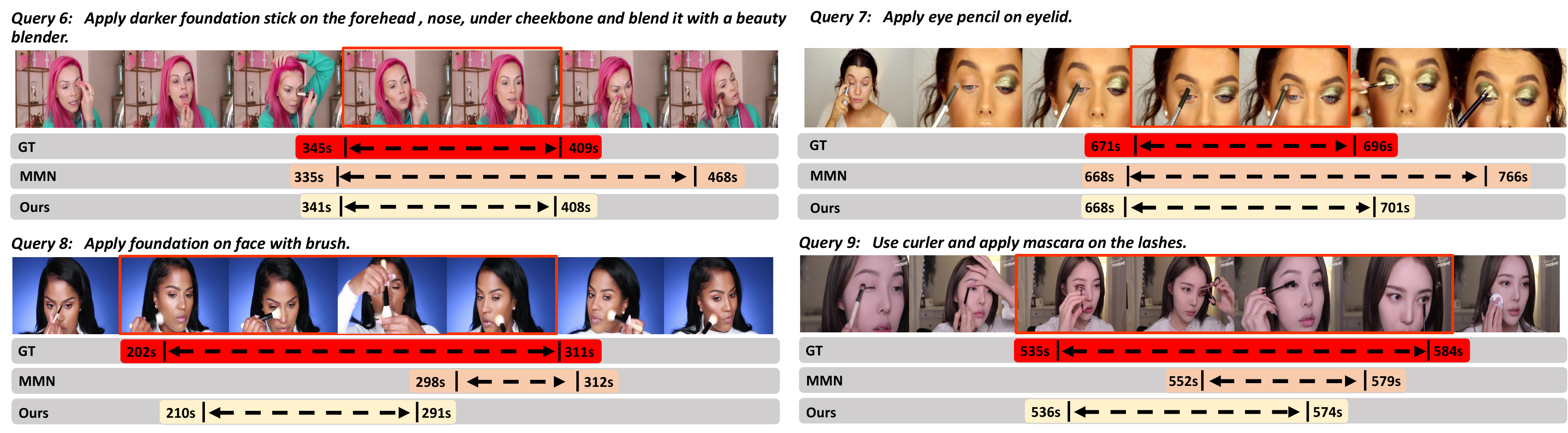}
\caption{Qualitative results of our approach (DPTMO) and the baseline (MMN) on the YouMakeup dataset. GT denotes the ground truth annotation, MMN denotes the prediction result of MMN, and ours denotes the result of our proposed method.}
\label{fig:visual}
\end{figure*}

\subsubsection{Visualization of dual-path temporal map optimization}\label{sec:visualize_dual_path}
To show temporal map optimization results of our proposed DPTMO model more intuitively, we further visualize some examples of query-agnostic and query-conditioned maps in the YouMakeup dataset. 
As shown in Fig.~\ref{fig:map_cmp}, in each example, we display the prediction results, and the 2D temporal candidates scores of ground truth, query-conditioned map, query-agnostic map, and joint prediction map. 
In addition, we also amplified the surrounding regions of ground truth on the right of each candidates map. 
In particular, we mark the interference candidates and corresponding coordinates in the query-agnostic map and the query-conditioned map with the arrows. 
The interference proposals in the same types of temporal maps are marked with the same color arrows. 

From the above visualization maps, we conclude some observations as follows: 
(1) The proposed DPTMO enjoys the merit of fine-grained semantic comprehension. 
Makeup video usually contains similar actions with subtle changes, making the prediction results noisier. 
In $Query$ 3, the instructional actions in this video are very similar. 
Specifically, each action shares the same queried activity \textit{``apply [something] on lips''}. 
Different actions can only be distinguished by the color or shape of cosmetics. 
Only the candidates with high confidence in both of the maps can be selected as the output, which effectively improve the fine-grained comprehension of our model. 
(2) Query-agnostic map and query-conditioned map collaboratively contribute to the joint prediction, both of which play an essential role in the proposed model. 
In $Queries$ 3 and 5, DPTMO achieves poor prediction in the query-conditioned map while better prediction in the query-conditioned map. 
From our observation, the poor prediction result is caused by some interference candidates. 
For $Query$ 3, the interference candidates in both the query-conditioned and query-agnostic maps are ``Apply eyeshadow on the eyes or eyelids with brush''. These actions are very similar to the query ``Apply concealer under the eyes with brush and fingers''.  
The same conclusion also can be found in $Query$ 2. 
In $Query$ 4, DPTMO achieves poor predictions in the query-agnostic map. In addition, there are some interference proposals in the query-agnostic map. We select one and mark the corresponding coordinates (\ie, 105s to 136s) below the video sequence. This corresponding action is ``Apply eyeshadow on the eyelids with fingers'', which is very similar to the query ``Apply blush on the cheek with fingers''. 
We can see that the joint predicted map is clearer than both query-conditioned and query-agnostic maps and closer to the ground truth. 
We can conclude that our dual-path temporal map optimization can correct poor prediction and reduce the noise from interference candidates. 
The above qualitative results also prove that the DPTMO model is more stable and accurate than single-path models.

\subsubsection{Visualization of Prediction Results}
To demonstrate the fined-grained semantic comprehension of the proposed model, we visualize some cases in Fig.~\ref{fig:visual}, which illustrate the prediction results of our model and MMN. 
For $Query$ 6 \textit{``Apply darker foundation stick on the forehead, nose, under cheekbone and blend it with a beauty blender''}, the frames in this video are of great similarity. 
In addition, this video contains multiple similar actions, which are continuous. 
The tiny difference only can be distinguished by the color of foundation sticks: ground truth describing \textit{``the darker foundation stick''} while the remain of MMN prediction is \textit{``lighter''} one. 
Similarly, for $Query$ 7, it is hard to make a clear distinction between \textit{``eyeliner''} and \textit{``eye shadow''}. \textit{``eyeliner''} and \textit{``eye shadow''} can only be distinguished by the colors and shapes. 
Additionally, the two makeup actions only occupy the tiny size of physical regions and are applied to the same location, requiring models to possess qualified fined-grained semantic comprehension. 
For above two queries, our model is able to locate the target action accurately. 
In $Query$ 8 \textit{``Apply foundation on face with brush''}, we find that in the interval of 263 seconds to 301 seconds in the video, the blogger only introduced cosmetics, without any action. 
This is the reason why our model and MMN localize totally different parts of the target segment. 
Both of them failed to notice the integrity of the target sequence, but our model prediction is more accurate. 
This results also indicate that our model can still make more accurate predictions under strong interference. 
For $Query$ 9 ``\textit{Use curler and apply mascara on the lashes}'', MMN only locates the later action ``\textit{apply mascara on the lashes}'' while our proposed method locates both ``\textit{use curler}'' and ``\textit{apply mascara}'' accurately. 
In summary, the proposed model is sensitive to fined-grained semantic information. 

\section{Conclusions}
In this paper, we focus on the task of video grounding in the specific make-up domain, and introduce a novel proposal-based framework named Dual-Path Temporal Map Optimization Network (DPTMO) with fine-grained semantic comprehension ability. 
The motivation of this paper is to eliminate interference candidates and allocate more attention to the target candidates. 
To achieve this, DPTMO models the inter-modal adjacent relations and queried-conditioned adjacent relations in two maps: a query-agnostic map and a query-conditioned map, respectively. 
Based on these maps, DPTMO makes the joint prediction, selecting the candidates with high scores in both maps as the output. 
Compared with previous 2D map methods, DPTMO effectively refines the quality of candidates and reduces noise in candidates map. 
Experimental results show that the proposed model brings significant improvements on the YouMakeup dataset. 
The dual-path temporal map optimization method introduced in this paper offers a new perspective for temporal map generation and evaluation in the challenging MTVG task. 
In the future, we plan to extract face landmarks for accurate face region recognition to improve grounding accuracy, and to use skeleton points combined with facial regions for more precise motion modeling. In addition, contrastive vision-language pre-training models can be employed to benefit the cross-modal matching process in this task.

\bibliographystyle{IEEEtran}
\bibliography{IEEEtrans.bib}

\end{document}